\documentclass[runningheads]{llncs}
\usepackage{graphicx}
\usepackage{comment}
\usepackage{amsmath,amssymb} 
\usepackage{color}

\usepackage{multirow}
\usepackage{subfigure}
\usepackage{algorithm}
\usepackage{algorithmic}
\usepackage{comment}
\usepackage{booktabs}
\usepackage{diagbox}
\usepackage{adjustbox}
\usepackage{mathrsfs}
\usepackage{appendix}
\usepackage[normalem]{ulem}
\useunder{\uline}{\ul}{}

\DeclareMathOperator*{\argmin}{arg\,min}

\usepackage[breaklinks=true,bookmarks=false]{hyperref}

\makeatletter
\def\thickhline{%
  \noalign{\ifnum0=`}\fi\hrule \@height \thickarrayrulewidth \futurelet
   \reserved@a\@xthickhline}
\def\@xthickhline{\ifx\reserved@a\thickhline
               \vskip\doublerulesep
               \vskip-\thickarrayrulewidth
             \fi
      \ifnum0=`{\fi}}
\makeatother

\newlength{\thickarrayrulewidth}
\begin{document}

\title{Rethinking the Distribution Gap of \\ Person Re-identification with \\ Camera-based Batch Normalization}


\title{Rethinking the Distribution Gap of \\ Person Re-identification with \\ Camera-based Batch Normalization}

\titlerunning{Rethinking the Distribution Gap of ReID with CBN}
\author{Zijie Zhuang\inst{1} \and Longhui Wei\inst{2,4}\and Lingxi Xie\inst{2} \and Tianyu Zhang\inst{2} \and Hengheng Zhang\inst{3} \and Haozhe Wu\inst{1} \and Haizhou Ai\inst{1} \and Qi Tian\inst{2}\\}

\authorrunning{Z.Zhuang et al.}

\institute{\textsuperscript{1}Tsinghua University\quad\textsuperscript{2}Huawei Inc.\\
\textsuperscript{3}Hefei University of Technology\quad\textsuperscript{4}University of Science and Technology of China\\
\email{\{jayzhuang42,weilh2568,198808xc,tianyu1949,imhmhm\}@gmail.com\\
wuhz1997@163.com, ahz@tsinghua.edu.cn, tian.qi1@huawei.com}}
\maketitle

\begin{abstract}
The fundamental difficulty in person re-identification (ReID) lies in learning the correspondence among individual cameras.
It strongly demands costly inter-camera annotations, yet the trained models are not guaranteed to transfer well to previously unseen cameras.
These problems significantly limit the application of ReID.
This paper rethinks the working mechanism of conventional ReID approaches and puts forward a new solution. 
With an effective operator named Camera-based Batch Normalization (CBN), we force the image data of all cameras to fall onto the same subspace, so that the distribution gap between any camera pair is largely shrunk. 
This alignment brings two benefits. 
First, the trained model enjoys better abilities to generalize across scenarios with unseen cameras as well as transfer across multiple training sets.
Second, we can rely on intra-camera annotations, which have been undervalued before due to the lack of cross-camera information, to achieve competitive ReID performance. 
Experiments on a wide range of ReID tasks demonstrate the effectiveness of our approach.
The code is available at https://github.com/automan000/Camera-based-Person-ReID.

\end{abstract}

\section{Introduction}
\label{sec:introduction}

Person re-identification (ReID) aims at matching identities across disjoint cameras.
Generally, it is achieved by mapping images from the same and different cameras into a feature space, where features of the same identity are closer than those of different identities. 
To learn the relations between identities from all cameras, there are two different objectives: learning the relations between identities in the same camera and learning identity relations across cameras.

\begin{figure}[t]
\centering
\subfigure[]{
\label{fig:show_img_diff}
\includegraphics[width=0.235\textwidth]{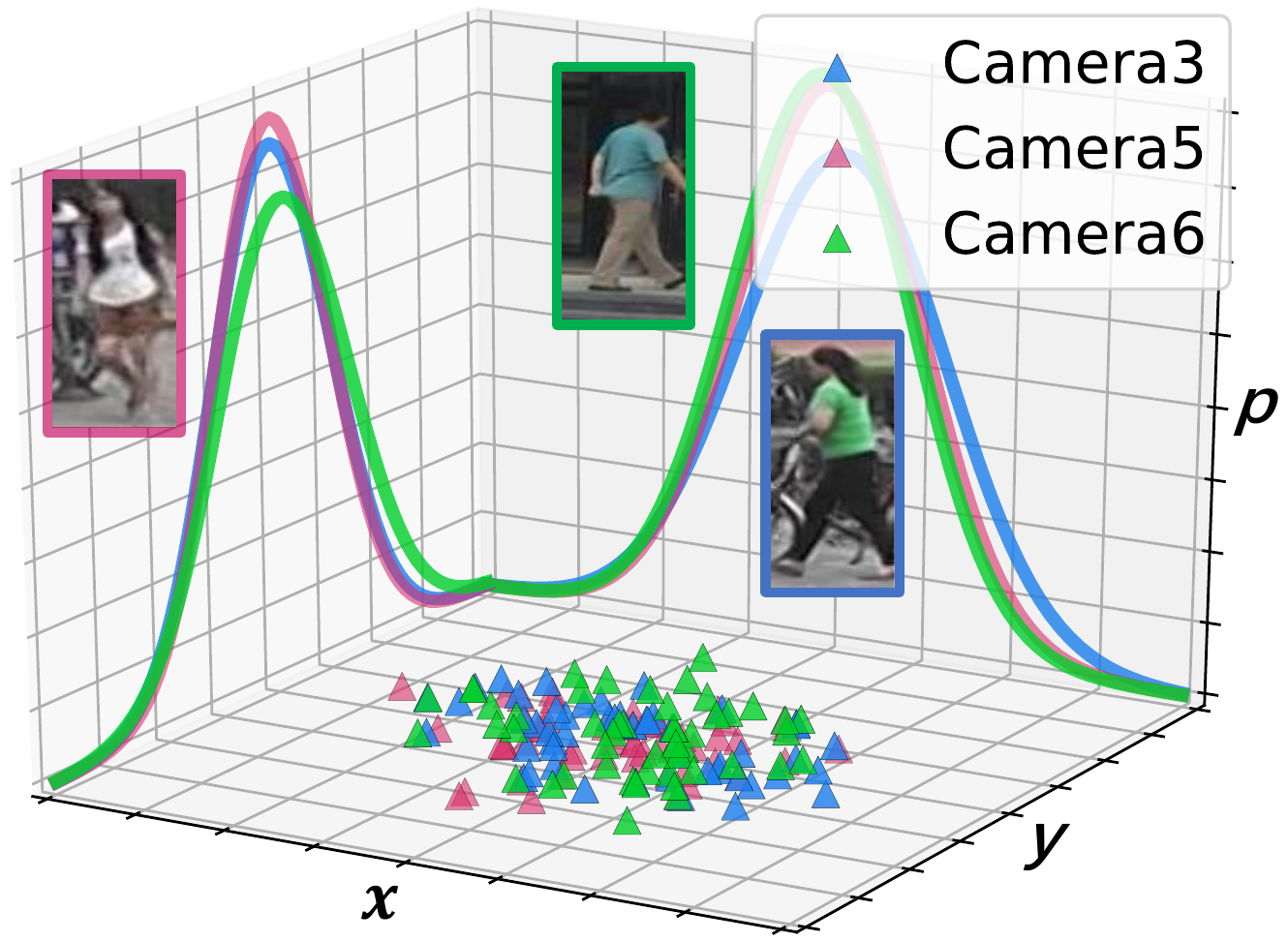}}
\subfigure[]{
\label{fig:dataset_bias}
\includegraphics[width=0.235\textwidth]{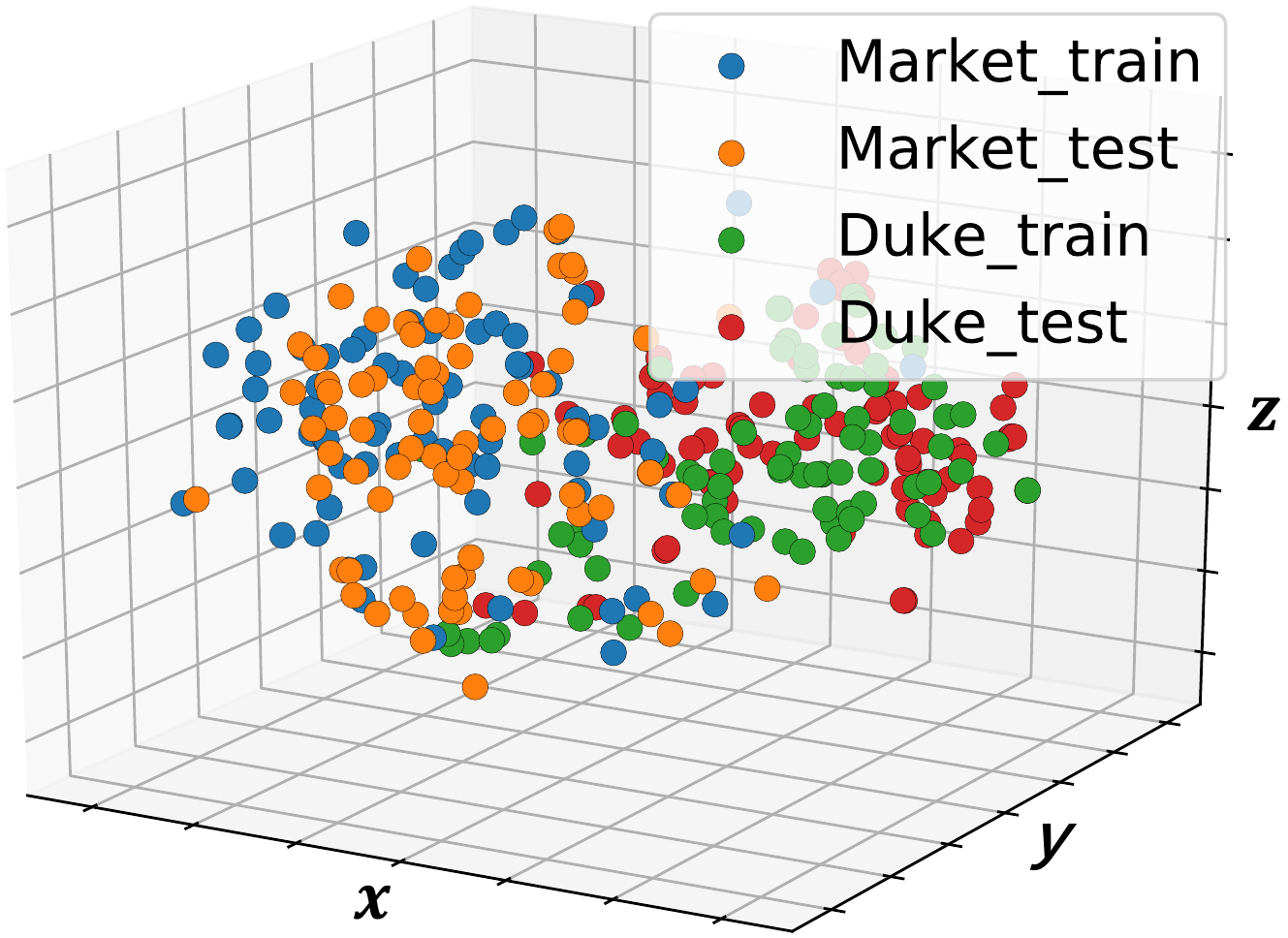}}
\subfigure[]{
\label{fig:visualize_formulation}
\includegraphics[width=0.39\textwidth]{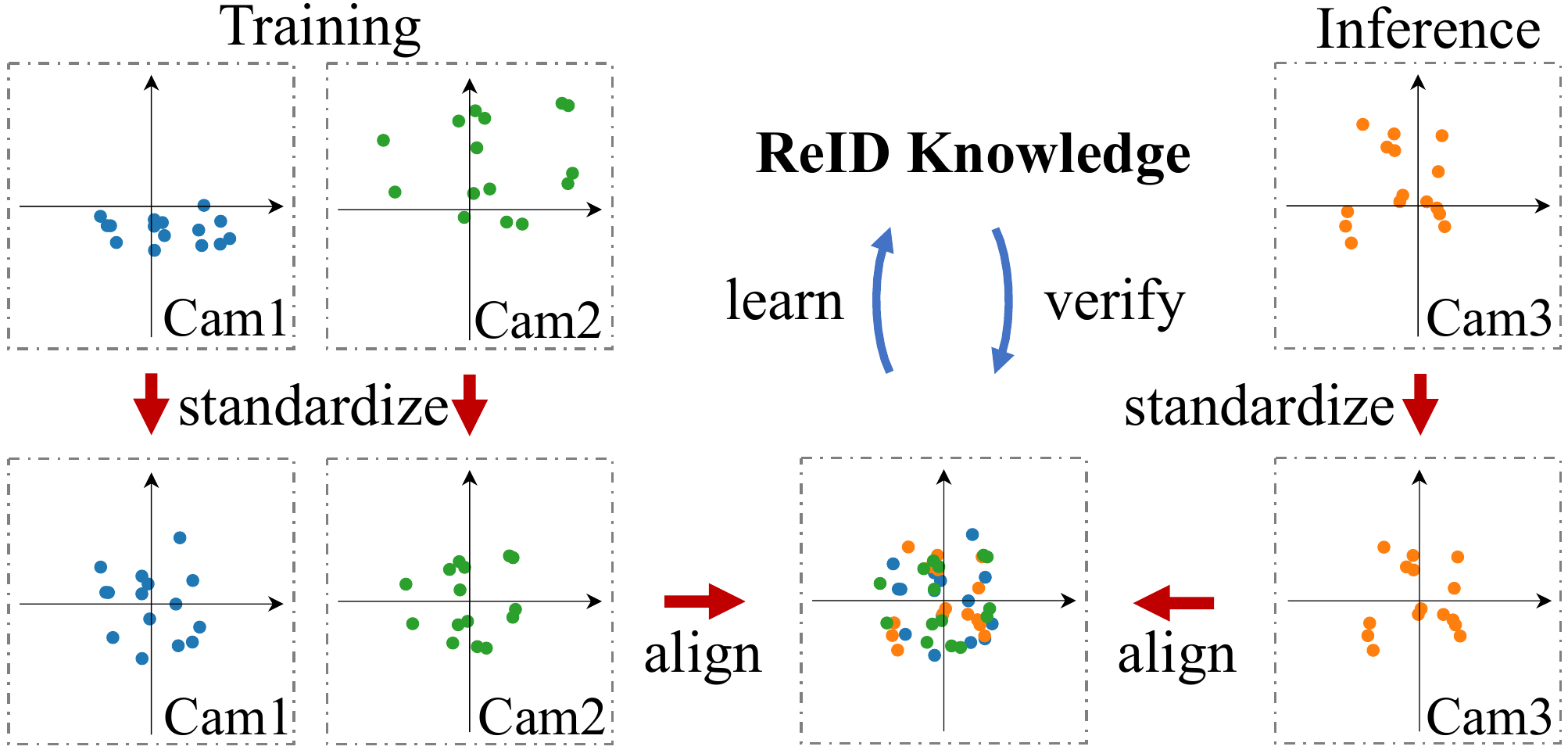}}
\caption{
(a) We visualize the distributions of several cameras in Market-1501.
Each curve corresponds to an approximated marginal density function.
Curves of different cameras demonstrate the differences between the corresponding distributions.
(b) The Barnes-Hut t-SNE~\cite{TSNE} visualization of the distribution inconsistency among datasets.
(c) Illustration of the proposed camera-based formulation.
Note that \textbf{Cam1}, \textbf{Cam2}, and \textbf{Cam3} could come from any ReID datasets.
This figure is best viewed in color.
}
\label{fig:visualize_inconsistency}
\end{figure}

However, there is an inconsistency between these two objectives. 
As shown in Fig.~\ref{fig:show_img_diff}, due to the large appearance variation caused by illumination conditions, camera views, {\em etc.}, images from different cameras are subject to distinct distributions.
Handling the distribution gap between cameras is crucial for inter-camera identity matching, yet learning within a single camera is much easier.
As a consequence, the conventional ReID approaches mainly focus on associating different cameras, which demands costly inter-camera annotations.
Besides, after learning on a training set, part of the learned knowledge is strongly correlated to the connections among these particular cameras, making the model generalize poorly on scenarios consisting of unseen cameras.
As shown in Fig.~\ref{fig:dataset_bias}, the ReID model learned on one dataset often has a limited ability of describing images from other datasets, \textit{i.e.}, its generalization ability across datasets is limited.
For simplicity, we denote this formulation neglecting within-dataset inconsistencies as the \textbf{dataset-based formulation}.
\textbf{}
We emphasize that lacking the ability to bridge the distribution gap between all cameras from all datasets leads to two problems: the unsatisfying generalization ability and the excessive dependence on inter-camera annotations.
To tackle these problems simultaneously, we propose to align the distribution of all cameras explicitly.
As shown in Fig.~\ref{fig:visualize_formulation}, we eliminate the distribution inconsistency between all cameras, so the ReID knowledge can always be learned, accumulated, and verified in the same input distribution, which facilitates the generalization ability across different ReID scenarios.
Moreover, with the aligned distributions among all cameras, intra- and inter-camera annotations can be regarded as the same, {\em i.e.}, labeling the image relations under the same input distribution.
This allows us to approximate the effect of inter-camera annotations with only intra-camera annotations.
It may relieve the exhaustive human labor for the costly inter-camera annotations.

We denote our solution that disassembles ReID datasets and aligns each camera independently as the \textbf{camera-based formulation}.
We implement it via an improved version of Batch Normalization (BN)~\cite{ioffe2015batch} named Camera-based Batch Normalization (CBN).
In training, CBN disassembles each mini-batch and standardizes the corresponding input according to its camera labels.
In testing, CBN utilizes few samples to approximate the BN statistics of every testing camera and standardizes the input to the training distribution.
In practice, multiple ReID tasks benefit from our work, such as \textit{fully-supervised learning}~\cite{Almazan2018,zheng2016a,SVDNET,RERANK,RANDOMERASING,Zhou2017Efficient}, \textit{direct transfer}~\cite{luo2019bag,huang2018eanet}, \textit{domain adaptation}~\cite{PTGAN,deng2018image,camstyle,PUL,song2018unsupervised,zheng2017unlabeled}, and \textit{incremental learning}~\cite{rannen2017encoder,li2018learning,kirkpatrick2017overcoming}.
Extensive experiments indicate that our method improves the performance of these tasks simultaneously, such as $0.9\%$, $5.7\%$, and $14.2\%$ averaged Rank-1 accuracy improvements on \textit{fully-supervised learning}, \textit{domain adaptation}, and \textit{direct transfer}, respectively, and $9.7\%$ less forgetting on Rank-1 accuracy for \textit{incremental learning}. 
Last but not least, even without inter-camera annotations, a \textit{weakly-supervised} pipeline~\cite{zhu2019intra} with our formulation can achieve competitive performance on multiple ReID datasets, which demonstrates that the value of intra-camera annotations may have been undervalued in the previous literature.
To conclude, our contribution is three-fold:
\begin{itemize}
\item In this paper, we emphasize the importance of aligning the distribution of all cameras and propose a camera-based formulation.
It can learn discriminative knowledge for ReID tasks while excluding training-set-specific information.
\item We implement our formulation with Camera-based Batch Normalization.
It facilitates the generalization and transfer ability of ReID models across different scenarios and makes better use of intra-camera annotations.
It provides a new solution for ReID tasks without costly inter-camera annotations.
\item Experiments on \textit{fully-supervised}, \textit{weakly-supervised}, \textit{direct transfer}, \textit{domain adaptation}, and \textit{incremental learning} tasks validate our method, which confirms the universality and effectiveness of our camera-based formulation.
\end{itemize}

\section{Related Work}
\label{sec:related}

Our formulation aligns the distribution per camera.
In training, it eliminates the distribution gap between all cameras.
ReID models can treat both intra-camera and inter-camera annotations equally and make better use of them, which benefits both \textit{fully-supervised} and \textit{weakly-supervised} ReID tasks.
It also guarantees that the distribution of each testing camera is aligned to the same training distribution.
Thus, the knowledge can better generalize and transfer across datasets.
It helps \textit{direct transfer}, \textit{domain adaptation}, and \textit{incremental learning}.
In this section, we briefly categorize and summarize previous works on the above ReID topics.

\noindent
\textbf{Supervision.}
The supervision in ReID tasks is usually in the form of identity annotations.
Although there are many outstanding unsupervised methods~\cite{yu2017cross,wu2019unsupervised,yu2019unsupervised,yu2018unsupervised} that do not need annotations, it is usually hard for them to achieve competitive performance as the supervised ReID methods.
For better performance, lots of previous methods~\cite{Almazan2018,zheng2016a,SVDNET,RERANK,RANDOMERASING,Zhou2017Efficient,kalayeh2018human,glad} utilized \textit{fully-supervised learning}, in which identity labels are annotated manually across all training cameras.  
Many of them designed spatial alignment~\cite{alignedreid,Sun2018_beyond,Suh_2018_ECCV}, visual attention ~\cite{han,CAN}, and semantic segmentation~\cite{kalayeh2018human,tian2018eliminating,MASKCAREID} for extracting accurate and fine-grained features.
GAN-based methods~\cite{liu2018pose,jiao2018deep,mao2019resolution} were also utilized for data augmentation.
However, although these methods achieved remarkable performance on ReID tasks, they required costly inter-camera annotations.
To reduce the cost of human labor, ReID researchers began to investigate~\textit{weakly-supervised learning}.
SCT~\cite{zhang2019single} presumes that each identity appears in only one camera.
In ICS~\cite{zhu2019intra}, an intra-camera supervision task is studied in which an identity could have different labels under different cameras.
In~\cite{lin2019bottom,lin2020unsupervised}, pseudo labels are used to supervised the ReID model.

\noindent
\textbf{Generalization.}
The generalization ability in ReID tasks denotes how well a trained model functions on unseen datasets, which is usually examined by \textit{direct transfer} tasks.
Researchers found that many fully-supervised ReID models perform poorly on unseen datasets~\cite{song2019generalizable,PTGAN,deng2018image}. 
To improve the generalization ability, various strategies were adopted as additional constraints to avoid over-fitting, such as label smoothing~\cite{luo2019bag} and sophisticated part alignment approaches~\cite{huang2018eanet}.

\noindent
\textbf{Transfer.}
The transfer ability in ReID tasks corresponds to the capability of ReID models transferring and preserving the discriminative knowledge across multiple training sets.
There are two related tasks.
\textit{Domain adaptation} transfers knowledge from labeled source domains to unlabeled target domains.
One solution~\cite{PTGAN,deng2018image,camstyle} bridged the domain gap by transferring source images to the target image style.
Other solutions~\cite{ssg,TJAIDL,PUL,MMFA,song2018unsupervised} utilized the knowledge learned from the source domain to mine the identity relations in target domains.
\textit{Incremental learning}~\cite{rannen2017encoder,li2018learning,kirkpatrick2017overcoming} also values the transfer ability.
Its goal is to preserve the previous knowledge and accumulate the common knowledge for all seen datasets.
A recent ReID work that relates to incremental learning is MASDF~\cite{incremental_reid}, which distilled and incorporated the knowledge from multiple datasets.

\section{Methodology}

\subsection{Conventional ReID: Learning Camera-related Knowledge}
\label{sec:flaws_in_conventional_formulation}

ReID is a task of retrieving identities according to their appearance.
Given a training set consisting of disjoint cameras, learning a ReID model on it requires two types of annotations: inter-camera annotations and intra-camera annotations.
The conventional ReID formulation regards a ReID dataset as a whole and learns the relations between identities as well as the connections between training cameras.
Given an image $\mathbf{I}^{\mathcal{D}_j}_i$ from any training set $\mathcal{D}_j$, the training goal of this formulation is:
\begin{equation}
{\argmin\mathbb{E}\!\left[\mathbf{y}_{i}^{\mathcal{D}_{j}}-\mathbf{g}^{\mathcal{D}_{j}}\left(\mathbf{f}^{\mathcal{D}_{j}}\!\left(\mathbf{I}^{\mathcal{D}_{j}}_{i}\right)\right)\right]},{\left(\mathbf{I}^{\mathcal{D}_{j}}_{i},\mathbf{y}_{i}^{\mathcal{D}_{j}}\right)}\in{\mathcal{D}_{j}},
\label{eq:old_formulation}
\end{equation}
where $\mathbf{f}^{\mathcal{D}_{j}}\left(\cdot\right)$ and $\mathbf{g}^{\mathcal{D}_{j}}\left(\cdot\right)$ are the corresponding feature extractor and classifier for $\mathcal{D}_{j}$, respectively.
$\mathbf{y}^{\mathcal{D}_{j}}_{i}$ denotes the identity label of the image $\mathbf{I}^{\mathcal{D}_{j}}_{i}$.

In our opinion, this formulation has three drawbacks.
First, images from different cameras, even of the same identity, are subject to distinct distributions.
To associate images across cameras, conventional approaches strongly demand the costly inter-camera annotations.
Meanwhile, the intra-camera annotations are less exploited since they provide little information across cameras.
Second, such learned knowledge not only discriminates the identities in the training set but also encodes the connections between training cameras.
These connections are associated with the particular training cameras and hard to generalize to other cameras, since the corresponding knowledge may not apply to the distribution of previously unseen cameras.
For example, when transferring a ReID model trained on Market-1501 to DukeMTMC-reID, it produces a poor Rank-1 accuracy of $37.0\%$ without fine-tuning.
Third, the learned knowledge is hard to preserve when being fine-tuned.
For instance, after fine-tuning the aforementioned model on DukeMTMC-reID, the Rank-1 accuracy drops $14.2\%$ on Market-1501, because it turns to fit the relations between the cameras in DukeMTMC-reID.
We analyze these three problems and find that the particular relations between training cameras are the primary cause of them.
Thus, we believe that the conventional method of handling these camera-related relations may need a re-design.

\subsection{Our Insight: Towards Camera-independent ReID}

We rethink the relations between cameras.
More specifically, we believe that the exclusive knowledge for bridging the distribution gap between the particular training cameras should be suppressed during training.
Such knowledge is associated to the cameras in the training set and sacrifices the discriminative and generalization ability on unseen scenarios.

To this end, we propose to align the distribution of all cameras explicitly, so that the distribution gap between all cameras is eliminated, and much less camera-specific knowledge will be learned during training.
We denote this formulation as the \textbf{camera-based formulation}.
To align the distribution of each camera, we estimate the raw distribution of each camera and standardize images from each camera with the corresponding distribution statistics.
We use $\mathbf{\boldsymbol{\eta}}\left(\cdot\right)$ to denote the estimated statistics related to the distribution of a camera. 
Then, given a related image $\mathbf{I}_{i}^{\left(c\right)}$, aligning the camera-wise distribution will transform this image as:
\begin{equation}
\tilde{\mathbf{I}}^{\left(c\right)}_{i} = \mathbf{DA}\left(\mathbf{I}_{i}^{\left(c\right)}; \boldsymbol{\eta}\left({c}\right)\right),
\end{equation}
where $\mathbf{DA}\left(\cdot\right)$ represents a distribution alignment mechanism, $\tilde{\mathbf{I}}_{i}^{\left(c\right)}$ denotes the aligned $\mathbf{I}_{i}^{\left(c\right)}$ and $\boldsymbol{\eta}\left({c}\right)$ is the estimated alignment parameters for camera $c$.
For any training set $\mathcal{D}_{j}$, we can now learn the ReID knowledge from this aligned distribution by replacing $\mathbf{I}^{\mathcal{D}_{j}}_{i}$ in Eq.~\ref{eq:old_formulation} with $\tilde{\mathbf{I}}^{\left(c\right)}_{i}$.

With the distributions of all cameras aligned by $\mathbf{DA}\left(\cdot\right)$, images from all these cameras can be regarded as distributing on a ``standardized camera''.
By learning on this ``standardized camera'', we eliminate the distribution gap between cameras, so the raw learning objectives within the same and across different cameras can be treated equally, making the training procedure more efficient and effective.
Besides, without the disturbance caused by the training-camera-related connections, the learned knowledge can generalize better across various ReID scenarios.
Last but not least, now that the additional knowledge for associating diverse distributions is much less required, our formulation can make better use of the intra-camera annotations.
It may relieve human labor for the costly inter-camera annotations, and provides a solution for ReID in a large-scale camera network with fewer demands of inter-camera annotations.

\subsection{Camera-based Batch Normalization}

In practice, a possible solution for aligning camera-related distributions is to conduct batch normalization in a camera-wise manner.
We propose the Camera-based Batch Normalization (CBN) for aligning the distribution of all training and testing cameras.
It is modified from the conventional Batch Normalization~\cite{ioffe2015batch}, and estimates camera-related statistics rather than dataset-related statistics. 

\noindent
\textbf{Batch Normalization Revisited.}
The Batch Normalization~\cite{ioffe2015batch} is designed to reduce the internal covariate shifting.
In training, it standardizes the data with the mini-batch statistics and records them for approximating the global statistics.
During testing, given an input $\mathbf{x}_{i}$, the output of the BN layer is:
\begin{equation}
\mathbf{\hat{x}}_{i} = \mathbf{\gamma} \frac{\mathbf{x}_{i} - \mathbf{\hat{\mu}}}{\sqrt{\mathbf{\hat{\sigma}}^\mathrm{2} + \epsilon}} + \mathbf{\beta},
\end{equation}
where $\mathbf{x}_{i}$ is the input and $\mathbf{\hat{x}}_{i}$ is the corresponding output.
$\hat{\mu}$ and $\hat{\sigma}^\mathrm{2}$ are the global mean and variance of the training set.
$\gamma$ and $\beta$ are two parameters learned during training.
In ReID tasks, BN has significant limitations.
It assumes and requires that all testing images are subject to the same training distribution.
However, this assumption is satisfied only when the cameras in the testing set and training set are exactly the same.
Otherwise, the standardization fails.

\noindent
\textbf{Batch Normalization within Cameras.}
Our Camera-based Batch Normalization (CBN) aligns all training and testing cameras independently.
It guarantees an invariant input distribution for learning, accumulating, and verifying the ReID knowledge.
Given images or corresponding intermediate features $\mathbf{x}_{m}^{\left(c\right)}$ from camera ${c}$, CBN standardizes them according to the camera-related statistics:

\begin{align}
\mu_{\left(c\right)} = \frac{1}{M}\sum_{m=1}^{M}\mathbf{x}_{{m}}^{\left(c\right)}, \ \
\sigma_{\left(c\right)}^{2} = \frac{1}{M}\sum_{m=1}^{M}\left(\mathbf{x}_{m}^{\left(c\right)} - \mu_{\left(c\right)}\right)^\mathrm{2}, \ \
\mathbf{\hat{x}}_{m} = & \gamma \frac{\mathbf{x}_{m} - \mu_{\left(c\right)}}{\sqrt{\sigma_{\left(c\right)}^\mathrm{2} + \epsilon}} + \beta ,
\end{align}
where $\mu_{\left(c\right)}$ and $\sigma_{\left(c\right)}^\mathrm{2}$ denote the mean and variance related to this camera ${c}$.
During training, we disassemble each mini-batch and calculate the camera-related mean and variance for each involved camera.
The camera with only one sampled images is ignored.
During testing, before employing the learned ReID model to extract features, the above statistics have to be renewed for every testing camera. 
In short, we collect several unlabeled images and calculate the camera-related statistics per testing camera.
Then, we employ these statistics and the learned weights to generate the final features.

\begin{figure}[t]
\centering
\subfigure[]{
\label{fig:overall_framework}
\includegraphics[width=0.41\textwidth]{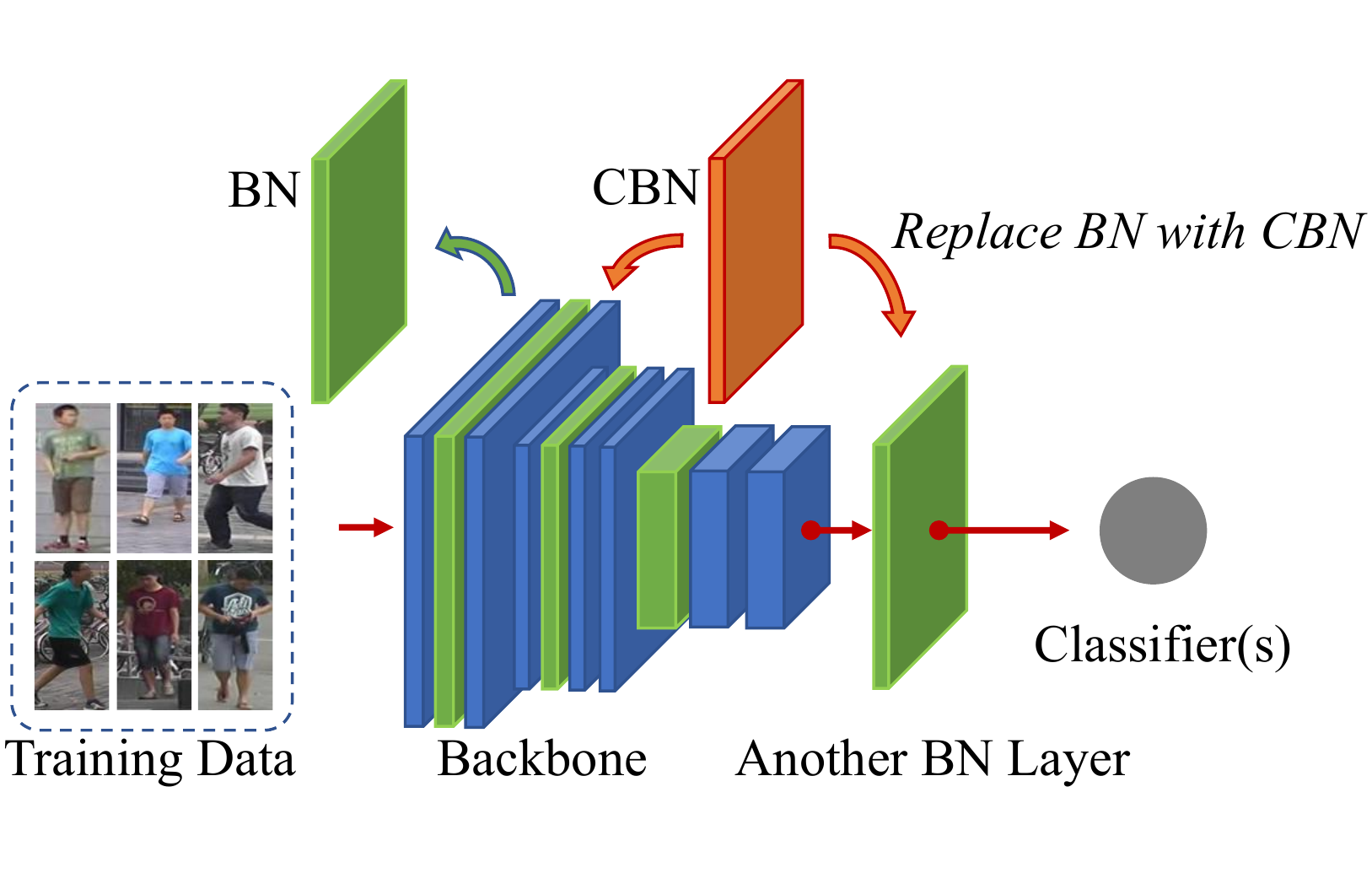}}
\subfigure[]{
\label{fig:data_free_incremental}
\includegraphics[width=0.26\textwidth]{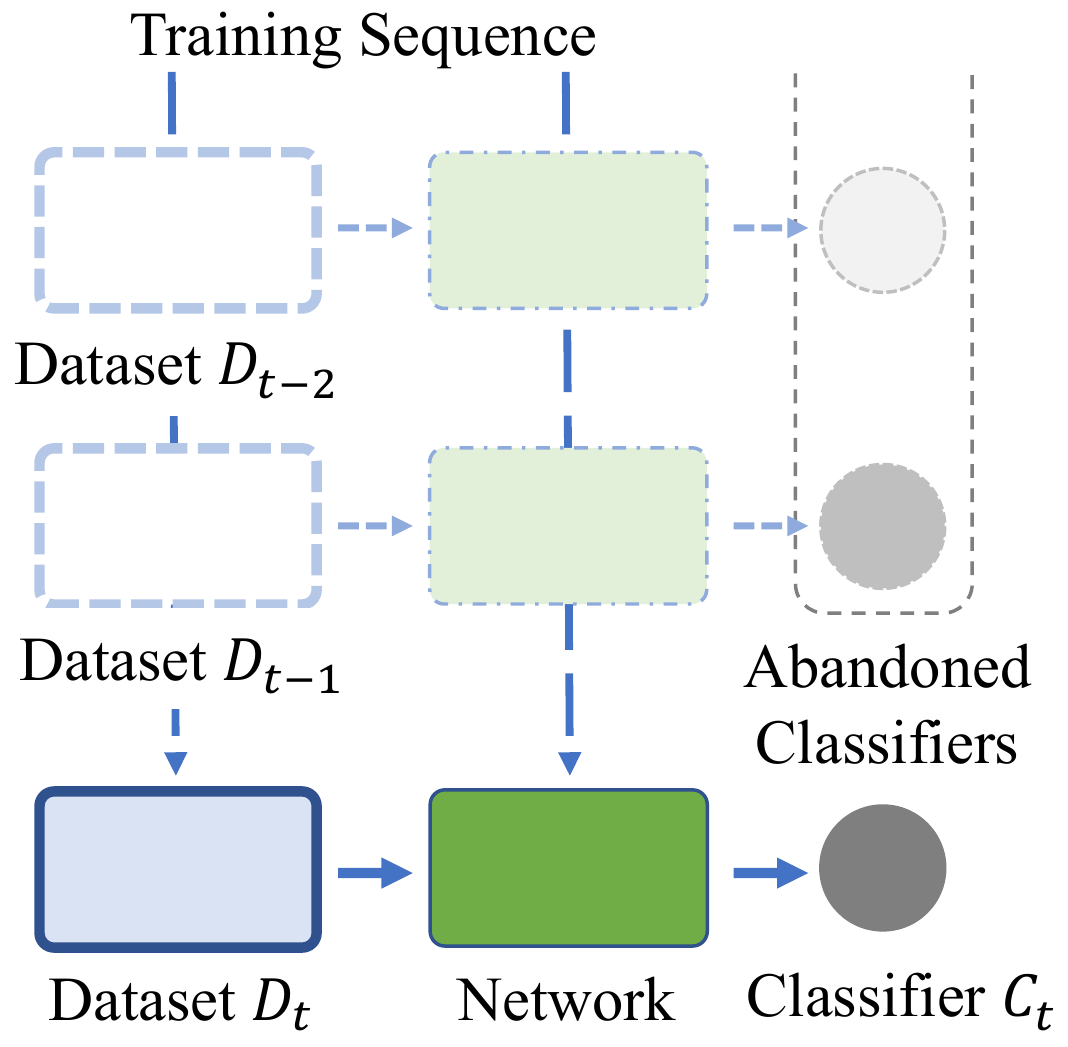}}
\subfigure[]{
\label{fig:memory_incremental}
\includegraphics[width=0.26\textwidth]{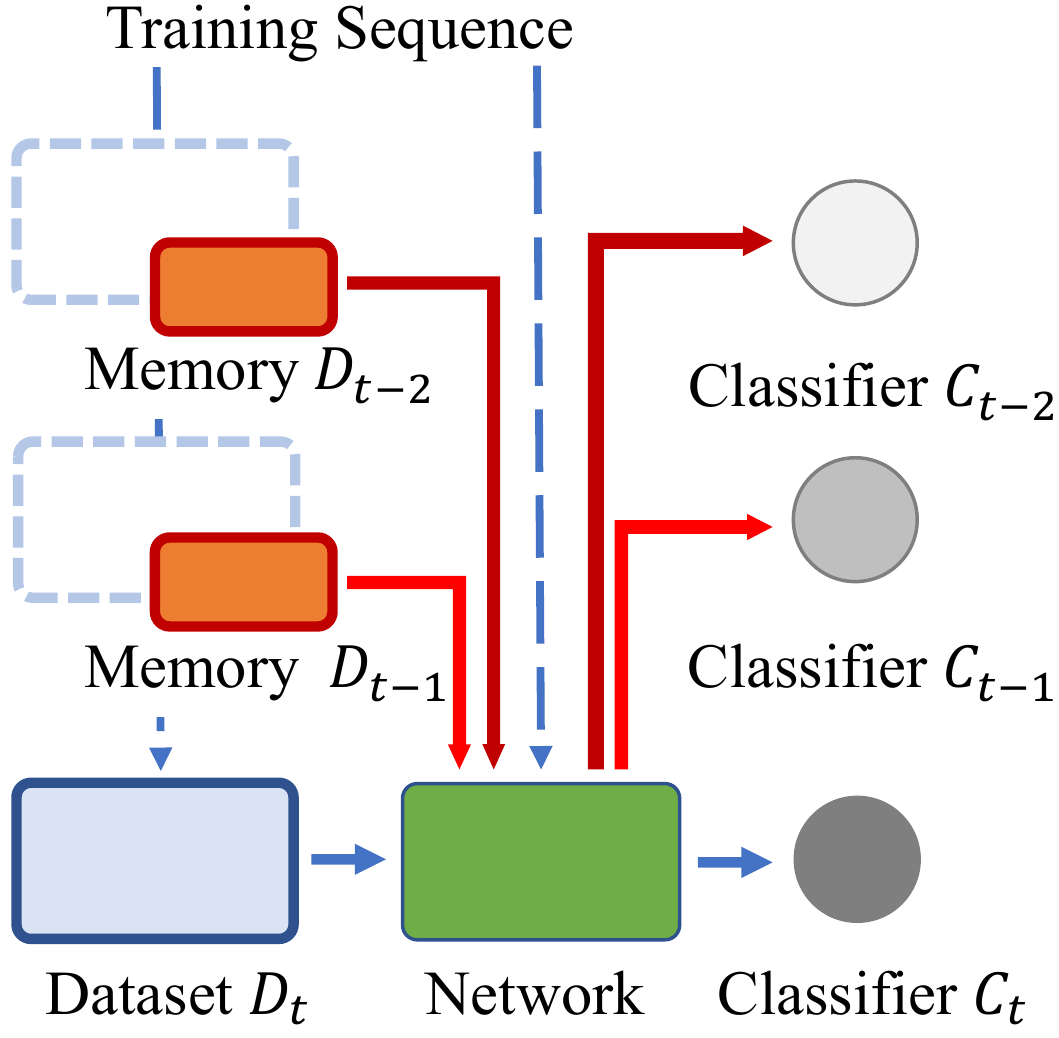}}
\caption{
Demonstrations of our bare-bones baseline network and two \textit{incremental learning} settings involved in this paper. 
(a) Given an arbitrary backbone with BN layers, we simply replace all BN layers with our CBN layers.
(b) \textbf{Data-Free}.
(c) \textbf{Replay}.
}
\label{fig:incremental_learning_settings}
\end{figure}

\subsection{Applying CBN to Multiple ReID Scenarios}
\label{sec:overall_framework}

The proposed CBN is generic and nearly cost-free for existing methods on multiple ReID tasks.
To demonstrate its superiority, we setup a bare-bones baseline, which only contains a deep neural network, an additional BN layer as the bottleneck, and a fully connected layer as the classifier.
As shown in Fig.~\ref{fig:overall_framework}, our camera-based formulation can be implemented by simply replacing all BN layers in a usual convolutional network with CBN layers.

With a modified network mentioned above, our camera-based formulation can be applied to many popular tasks, such as \textit{fully-supervised learning}, \textit{weakly-supervised learning}, \textit{direct transfer}, and \textit{domain adaptation}.
Apart from them, we also evaluate a rarely discussed ReID task, {\em i.e.}, \textit{incremental learning}.
It studies the problem of learning knowledge incrementally from a sequence of training sets while preserving and accumulating the previously learned knowledge.
As shown in Fig.~\ref{fig:incremental_learning_settings}, we propose two settings.
(1)~\textbf{Data-Free}: once we finish the training procedure on a dataset, the training data along with the corresponding classifier are abandoned.
When training the model on the subsequent training sets, the old data will never show up again.
(2)~\textbf{Replay}: unlike Data-Free, we construct an exemplar set from each old training set.
The exemplar set and the corresponding classifier are preserved and used during the entire training sequence.

\subsection{Discussions}
\label{sec:discussions}

\textbf{Bridging ReID Tasks.} 
We briefly demonstrate our understandings of the relations between ReID tasks and how we bridge these tasks.
Different ReID tasks handle different combinations of training and testing sets.
Since datasets have distinct cameras, previous methods have to learn exclusive relations between particular training cameras and adapt them to specific testing camera sets.
Our formulation aligns the distribution of all cameras for learning and testing ReID knowledge, and suppresses the exclusive training-camera relations.
It may reveal the latent connections between ReID tasks.
First, by aligning the distribution of seen and unseen cameras, \textit{fully-supervised learning} and \textit{direct transfer} are united since training and testing distributions are always aligned in a camera-wise manner.
Second, since there is no need to learn relations between distinct camera-related distributions, intra- and inter-camera annotations can be treated almost equally.
Knowledge is better shared among cameras which helps \textit{fully-} and \textit{weakly-supervised learning}.
Third, with the aligned training and testing distributions, it is more efficient to learn, accumulate, and preserve knowledge across datasets.
It offers an elegant solution to preserve old knowledge (\textit{incremental learning}) and absorb new knowledge (\textit{domain adaptation}) in the same model.

\noindent
\textbf{Relationship to Previous Works.} 
There are two types of previous works that closely relate to ours: camera-related methods and BN variants.
Same with our work, camera-related methods such as CamStyle~\cite{camstyle} and CAMEL~\cite{yu2017cross} noticed the camera view discrepancy inside the dataset. 
CamStyle augmented the dataset by transferring the image style in a camera-to-camera manner, but still learned ReID models in the dataset-based formulation.
Consequently, transferring across datasets is still difficult.
CAMEL~\cite{yu2017cross} is the most similar work with ours, which learned camera-related projections and mapped camera-related distributions into an implicit common distribution. 
However, these projections are associated with the particular training cameras, limiting its ability to transfer across datasets. 
BN variants such as AdaBN also inspire us. 
AdaBN aligned the distribution of the entire dataset.
It neither eliminated the camera-related relations in training, nor handled the camera-related distribution gap in testing. 
Unlike them, CBN is specially designed for our camera-based formulation.
It is much more general and precise for ReID tasks.
More comparisons and discussions will be provided in Secs.~\ref{sec:experiments} and~\ref{sec:ablation}.

\section{Experiments}

\subsection{Experiment Setup}

\textbf{Datasets.}
We utilize three large scale ReID datasets, including Market-1501~\cite{MARKET}, DukeMTMC-reID~\cite{zheng2017unlabeled}, and MSMT17~\cite{PTGAN}. 
Market-1501 dataset has $1\rm{,}501$ identities in total.
$751$ identities are used for training and the rest for testing.
The training set contains $12\rm{,}936$ images and the testing set contains $15\rm{,}913$ images.
DukeMTMC-reID dataset contains $16\rm{,}522$ images of $702$ identities for training, and $1\rm{,}110$ identities with $17\rm{,}661$ images are used for testing.
MSMT17 dataset is the current largest ReID dataset with $126\rm{,}441$ images of $4\rm{,}101$ identities from $15$ cameras.
For short, we denote Market-1501 as Market, DukeMTMC-reID as Duke, and MSMT17 as MSMT in the rest of this paper.
\textit{
It is worth noting that in these datasets, the training and testing subsets contain the same camera combinations.
It could be the reason that previous dataset-based methods create remarkable fully-supervised performance but catastrophic direct transfer results.}

\noindent
\textbf{Implementation Details.}
In this paper, all experiments are conducted with PyTorch.
In both training and testing, the image size is $256\times128$ and the batch size is $64$.
In training, we sample $4$ images for each identity.
The baseline network presented in Sec.~\ref{sec:overall_framework} uses the ResNet-50~\cite{ResNet} as the backbone.
To train this network, we adopt SGD optimizer with momentum~\cite{momentum} of $0.9$ and weight decay of $5\times10^{-4}$.
Moreover, the initial learning rate is $0.01$, and it decays after the $40th$ epoch by a factor of $10$.
For all experiments, the training stage will end up with $60$ epochs. 
For incremental learning, we include a warm-up stage.
In this stage, we freeze the backbone and only fine-tune the classifier(s) to avoid damaging the previously learned knowledge.
During testing, our framework will first sample a few unlabeled images from each camera and use them to approximate the camera-related statistics.
Then, these statistics are fixed and employed to process the corresponding testing images.
Following the conventions, mean Average Precision (mAP) and Cumulative Matching Characteristic (CMC) curves are utilized for evaluations.

\begin{table}[!t]
\centering
\caption{
Results of the baseline method with our formulation and the conventional formulation.
The fully-supervised learning results are in \textit{italics}.
}
\setlength{\tabcolsep}{1.7mm}{
\begin{tabular}{c|c|cc|cc|cc}
\thickhline
\multirow{2}{*}{{Training Set}} & {Testing Set} & \multicolumn{2}{c|}{{Market}}           & \multicolumn{2}{c|}{{Duke}}       & \multicolumn{2}{c}{{MSMT}}        \\  
\cline{2-8}       & {Formulation}      & {Rank-1}     & {mAP}        & {Rank-1}      & {mAP}        & {Rank-1}      & {mAP}        \\
\hline \hline
\multirow{2}{*}{{Market}}       & {Conventional}    & \textit{90.2}           & \textit{74.0}          & 37.0          & 20.7                & 17.1             & 5.5                 \\
                                       & {Ours}        & \textit{\textbf{91.3}}  & \textit{\textbf{77.3}} & \textbf{58.7}     & \textbf{38.2}       & \textbf{25.3}         & \textbf{9.5}        \\ \hline 
\multirow{2}{*}{{Duke}}         & {Conventional}    & 53.2            & 25.1                & \textit{81.5}          & \textit{66.6}          & 27.2                 & 9.1                 \\
                                       & {Ours}        & \textbf{72.7}      & \textbf{43.0}       & \textit{\textbf{82.5}}  & \textit{\textbf{67.3}} & \textbf{35.4}       & \textbf{13.0}         \\ \hline
\multirow{2}{*}{{MSMT}}         & {Conventional}    & 58.1                & 30.8                & 57.8               & 38.4                & \textit{71.5}             & \textit{42.3}          \\
                                       & {Ours}        & \textbf{73.7}         & \textbf{45.0}       & \textbf{66.2}          & \textbf{46.7}       & \textit{\textbf{72.8}}  & \textit{\textbf{42.9}} \\
\thickhline
\end{tabular}}
\label{tab:final_results_large_table}
\end{table}

\subsection{Performance on Different ReID Tasks}
\label{sec:experiments}

We evaluate our proposed method on five types of ReID tasks,~\emph{i.e.}, \textit{fully-supervised learning}, \textit{weakly-supervised learning}, \textit{direct transfer}, \textit{domain adaptation}, and \textit{incremental learning}.
The corresponding experiments are organized as follows.
First, we demonstrate the importance of aligning the distribution of all cameras from all datasets, and simultaneously conduct \textit{fully-supervised learning} and \textit{direct transfer} on multiple ReID datasets.
Second, we demonstrate that it is possible to learn discriminative knowledge with only intra-camera annotations.
We utilize the network architecture in Sec.~\ref{sec:overall_framework} to compare the \textit{fully-supervised learning} and \textit{weakly-supervised learning}.
To evaluate the generalization ability, \textit{direct transfer} is also conducted for these two settings.
Third, we evaluate the transfer ability of our method.
This part of experiments includes \textit{domain adaptation}, {\em i.e.}, transferring the knowledge from the old domain to new domains, and \textit{incremental learning}, {\em i.e.}, preserving the old knowledge and accumulating the common knowledge for all training sets.

Note that, for simplicity, we denote the results of training and testing the model on the same dataset with fully annotated data as the \textit{fully-supervised learning results}.
For similar experiments that only use the intra-camera annotations, we denote their results as the \textit{weakly-supervised learning results}.

\begin{table}[t]
\centering
\caption{
Results of the state-of-the-art fully-supervised learning methods.
BoT* denotes our results with the official BoT code.
In BoT*, Random Erasing is disabled due to its negative effect on direct transfer.
Unless otherwise stated, the \textbf{baseline} method in the following sections refers to the network described in Sec.~\ref{sec:overall_framework}.
}
\setlength{\tabcolsep}{0.9mm}{
\begin{tabular}{c|cccc|cccc}
\thickhline
\multirow{2}{*}{{Method}} & \multicolumn{4}{c|}{{Market}}   & \multicolumn{4}{c}{{Duke}}    \\  
\cline{2-9}   & {Rank-1} & {Rank-5} & {Rank-10} & {mAP}  & {Rank-1} & {Rank-5} & {Rank-10} & {mAP} \\
\hline \hline
{CamStyle}~\cite{camstyle}    & 88.1            & -               & -                & 68.7       & 75.3     & -         & -         & 53.5  \\
{MLFN}~\cite{mlfn}    & 90.0            & -               & -                & 74.3              & 81.0      & -       & -       & 62.8    \\
{SCPNet}~\cite{scpnet}               & 91.2            & 97.0            & -      & 75.2   & 80.3            & 89.6     & -       & 62.6    \\
{HA-CNN}~\cite{hacnn}                & 91.2            & -               & -      & 75.7       & 80.5     & -      & -     & 63.8     \\
{PGFA}~\cite{pgfa}                   & 91.2            & -               & -       & 76.8   & 82.6        & -      & -     & 65.5      \\
{MVP}~\cite{mvp}                   & 91.4            & -               & -                & 80.5      & 83.4      & -     & -      & 70.0   \\
{SGGNN}~\cite{sggnn}                &92.3             &96.1             &97.4               &82.8     &81.1       &88.4   &91.2      &68.2        \\
{SPReID}~\cite{kalayeh2018human}   & 92.5            & 97.2            & 98.1         & 81.3       & 84.4     & 91.9       & 93.7   & 71.0   \\
{BoT}*~\cite{luo2019bag}            & 93.6            & 97.6            & 98.4             & 82.2      & 84.3    & 91.9      & 94.2    & 70.1    \\
{PCB+RPP}~\cite{Sun2018_beyond}        & 93.8            & 97.5            & 98.5             & 81.6     & 83.3     & 90.5    & 92.5    & 69.2    \\ 
{OSNet}~\cite{osnet}        & 94.8            & -            & -             & 84.9     & 88.6     & -    & -    & 73.5    \\
{VA-reID}~\cite{vareid}        & \textbf{96.2}    & \textbf{98.7}    & -    & \textbf{91.7}     & \textbf{91.6}     & \textbf{96.2}    & -    & \textbf{84.5}    \\ \hline \hline
{Baseline}                     & 90.2      & 96.7      & 97.9      & 74.0    & 81.5      & 91.4         & 94.0       & 66.6     \\
\textbf{Ours+Baseline}            & 91.3            & 97.1       & 98.4      & 77.3     & 82.5      & 91.7       & 94.1       & 67.3    \\
\textbf{Ours+BoT*}    & 94.3   & 97.9   & 98.7    & 83.6 & 84.8   & 92.5   & 95.2    & 70.1  \\
\thickhline
\end{tabular}}
\label{tab:results_fully_supervised}
\end{table}

\noindent\textbf{Supervisions and Generalization.}
In this section, we evaluate and analyze the supervisions and the generalization ability in ReID tasks.
For all experiments in this section, the testing results on both the training domain and other unseen testing domains are always obtained by the same learned model.
We first conduct experiments on \textit{fully-supervised learning} and \textit{direct transfer}.
As shown in Tab.~\ref{tab:final_results_large_table}, our proposed method shows good advantages,~\emph{e.g.}, there is an averaged $1.1\%$ improvement in Rank-1 accuracy for the \textit{fully-supervised learning} task.
Meanwhile, without bells and whistles, there is an average $13.6\%$ improvement in Rank-1 accuracy for the \textit{direct transfer} task.
We recognize that our method has to collect a few unlabeled samples from each testing camera for estimating the camera-related statistics.
However, this process is fast and nearly cost-free.

Our method can also boost previous methods.
Take BoT~\cite{luo2019bag}, a recent state-of-the-art method, as an example.
We integrate our proposed CBN into BoT and conduct experiments with almost the same settings as in the original paper, including the network architecture, objective functions, and training strategies.
The only difference is that we disable Random Erasing~\cite{RANDOMERASING} due to its constant negative effects on \textit{direct transfer}.
The results of the \textit{fully-supervised learning} on Market and Duke are shown in Tab.~\ref{tab:results_fully_supervised}.
It should be pointed out that in \textit{fully-supervised learning}, training and testing subsets contain the same cameras.
Therefore, there is no significant shift among the BN statistics of the training set and the testing set, which favors the conventional formulation.
Even so, our method still improves the performance on both Market and Duke.
We believe that both aligning camera-wise distributions and better utilizing all annotations contribute to these improvements.
Moreover, we also present results on \textit{direct transfer} in Tab.~\ref{tab:results_domain_adaptation}.
It is clear that our method improves BoT significantly,~\emph{e.g.}, there is a $15.3\%$ Rank-1 improvement when training on Duke but testing on Market.
These improvements on both \textit{fully-supervised learning} and \textit{direct transfer} demonstrate the advantages of our camera-based formulation.

\begin{table}[!t]
\centering
\caption{
The comparisons of fully- and weakly-supervised learning.
Results of training and testing on the same domain are in \textit{italics}.
MT~\cite{zhu2019intra} is our baseline.
Except for the camera-based formulation, our weakly-supervised model follows all its settings.
}
\setlength{\tabcolsep}{1.8mm}{
\begin{tabular}{c|c|cc|cc|cc}
\thickhline
\multirow{2}{*}{{Training Set}} & {Testing Set} & \multicolumn{2}{c|}{{Market}}           & \multicolumn{2}{c|}{{Duke}}       & \multicolumn{2}{c}{{MSMT}}        \\  
\cline{2-8}       & {Supervision}      & {Rank-1}     & {mAP}        & {Rank-1}      & {mAP}        & {Rank-1}      & {mAP}        \\
\hline \hline
\multirow{3}{*}{{Market}}       & {MT~\cite{zhu2019intra}}    & \textit{78.4}           & \textit{52.1}          & $-$          & $-$                & $-$             & $-$                 \\
                                       & {Weakly}    & \textit{83.3}           & \textit{60.4}          & 48.9          & 29.7                & \textbf{26.8}             & \textbf{9.6}                 \\
                                       & {Fully}        & \textit{\textbf{91.3}}  & \textit{\textbf{77.3}} & \textbf{58.7}     & \textbf{38.2}       & 25.3         & 9.5        \\ \hline 
\multirow{3}{*}{{Duke}}         & {MT}    & $-$            & $-$                & \textit{65.2}          & \textit{44.7}          & $-$                 & $-$                \\
                                       & {Weakly}    & 68.4            & 37.7                & \textit{73.9}          & \textit{54.4}          & 33.7                 & 11.9                 \\
                                       & {Fully}        & \textbf{72.7}      & \textbf{43.0}       & \textit{\textbf{82.5}}  & \textit{\textbf{67.3}} & \textbf{35.4}       & \textbf{13.0}         \\ \hline
\multirow{3}{*}{{MSMT}}         & {MT}    & $-$         & $-$                 & $-$              & $-$               & \textit{39.6}             & \textit{15.9}          \\
                                       & {Weakly}    & 68.3                & 37.2                & 59.2               & 38.2                & \textit{49.4}             & \textit{21.5}          \\
                                       & {Fully}        & \textbf{73.7}         & \textbf{45.0}       & \textbf{66.2}          & \textbf{46.7}       & \textit{\textbf{72.8}}  & \textit{\textbf{42.9}} \\
\thickhline
\end{tabular}}
\label{tab:compare_to_weakly_supervised}
\end{table}

\noindent\textbf{Weak Supervisions.}
As we demonstrated in Sec.~\ref{sec:flaws_in_conventional_formulation}, the conventional ReID formulation strongly demands the inter-camera annotations for associating identities under distinct camera-related distributions.
Since our method eliminates the distribution gap between cameras, the intra-camera annotations can be better used for learning the appearance features.
We compare the performance of using all annotations (\textit{fully-supervised learning}) and only intra-camera annotations (\textit{weakly-supervised learning}).
The results are in Tab.~\ref{tab:compare_to_weakly_supervised}.
For weakly-supervised experiments, we follow the same settings in MT~\cite{zhu2019intra}.
Since there are no inter-camera annotations, the identity labels of different cameras are independent, and we assign each individual camera with a separate classifier.
Each of these classifiers is supervised by the corresponding intra-camera identity labels.  
Surprisingly, even without inter-camera annotations, the \textit{weakly-supervised learning} achieves competitive performance.
According to these results, we believe that the importance of intra-camera annotations is significantly undervalued.

\noindent\textbf{Transfer.}
In this section, we evaluate the ability to transfer ReID knowledge between the old and new datasets.
First, we evaluate the ability to transfer previous knowledge to new domains.
The related task is \textit{domain adaptation}, which usually involves a labeled source training set and another unlabeled target training set.
We integrate our formulation into a recent state-of-the-art method ECN~\cite{ECN}.
The results are shown in Tab.~\ref{tab:results_domain_adaptation}.
By aligning the distributions of source labeled images and target unlabeled images, the performance of ECN is largely boosted,~\emph{e.g.}, when transferring from Duke to Market, the Rank-1 accuracy and mAP are improved by $6.6\%$ and $9.0\%$, respectively.
Meanwhile, compared to other methods that also utilize camera labels, such as CamStyle~\cite{camstyle} and CASCL~\cite{wu2019unsupervised}, our method outperforms them significantly.
These improvements demonstrate the effectiveness of our camera-based formulation in \textit{domain adaptation}.

\begin{table}[t]
\caption{
The results of testing ReID models across datasets.
$\ddagger$ marks methods that only use the source domain data for training, {\em i.e.}, direct transfer. 
Other methods listed in this table utilize both the source and target training data, {\em i.e.}, domain adaptation. 
}
\setlength{\tabcolsep}{0.9mm}{
\begin{tabular}{c|cccc|cccc}
\thickhline
\multirow{2}{*}{{Method}} & \multicolumn{4}{c|}{{Duke to Market}}           & \multicolumn{4}{c}{{Market to Duke}}                     \\ 
\cline{2-9}                   & {Rank-1} & {Rank-5} & {Rank-10} & {mAP}  & {Rank-1} & {Rank-5} & {Rank-10} & {mAP} \\ \hline \hline
\multicolumn{1}{c|}{{UMDL}~\cite{UMDL}}                        & 34.5     & 52.6       & 59.6        & 12.4        & 18.5            & 31.4            & 37.6             & 7.3 \\ 
\multicolumn{1}{c|}{{PTGAN}~\cite{PTGAN}}                      & 38.6     & -          & 66.1         & -           & 27.4            & -               & 50.7             & -  \\ 
\multicolumn{1}{c|}{{PUL}~\cite{PUL}}                          & 45.5     & 60.7       & 66.7        & 20.5        & 30.0            & 43.4            & 48.5             & 16.4 \\ 
\multicolumn{1}{c|}{{SPGAN}~\cite{deng2018image}}              & 51.5     & 70.1       & 76.8         & 22.8         & 41.1            & 56.6            & 63.0             & 22.3  \\ 
\multicolumn{1}{c|}{{BoT*$^\ddagger$}~\cite{luo2019bag}}       & 53.3     & 69.7       & 76.4         & 24.9       & 43.9            & 58.8            & 64.9             & 26.1  \\ 
\multicolumn{1}{c|}{{MMFA}~\cite{MMFA}}                        & 56.7     & 75.0       & 81.8            & 27.4         & 45.3            & 59.8            & 66.3             & 24.7  \\ 
\multicolumn{1}{c|}{{TJ-AIDL}~\cite{TJAIDL}}                   & 58.2     & 74.8       & 81.1           & 26.5         & 44.3            & 59.6            & 65.0             & 23.0  \\ 
\multicolumn{1}{c|}{{CamStyle}~\cite{camstyle}}                & 58.8     & 78.2       & 84.3          & 27.4          & 48.4            & 62.5            & 68.9             & 25.1 \\ 
\multicolumn{1}{c|}{{HHL}~\cite{HHL}}                          & 62.2     & 78.8       & 84.0          & 31.4         & 46.9            & 61.0            & 66.7             & 27.2  \\
\multicolumn{1}{c|}{{CASCL}~\cite{wu2019unsupervised}}         & 64.7     & 80.2       & 85.6          & 35.6          & 51.5            & 66.7            & 71.7             & 30.5 \\
\multicolumn{1}{c|}{{ECN}~\cite{ECN}}                          & 75.1     & 87.6       & 91.6          & 43.0         & 63.3            & 75.8            & 80.4             & 40.4  \\ \hline \hline
\multicolumn{1}{c|}{{Baseline$^\ddagger$}}                     & 53.2     & 70.0       & 76.0          & 25.1         & 37.0            & 52.6            & 58.9             & 20.7  \\
\multicolumn{1}{c|}{\textbf{Ours+BoT*$^\ddagger$}}                    & 68.6     & 82.5       & 87.7         & 39.0          & 60.6            & 74.0            & 78.5             & 39.8 \\ 
\multicolumn{1}{c|}{\textbf{Ours+Baseline$^\ddagger$}}                & 72.7     & 85.8       & 90.7        & 43.0         & 58.7            & 74.1            & 78.1             & 38.2  \\ 
\multicolumn{1}{c|}{\textbf{Ours+ECN}}         & \textbf{81.7}     & \textbf{91.9}   & \textbf{94.7}    & \textbf{52.0}       & \textbf{68.0}    & \textbf{80.0}    & \textbf{83.9}    & \textbf{44.9}\\
\thickhline
\end{tabular}}
\label{tab:results_domain_adaptation}
\end{table}

Second, we evaluate the ability to preserve old knowledge as well as accumulate common knowledge for all seen datasets when being fine-tuned.
\textit{Incremental learning}, which fine-tunes a model on a sequence of training sets, is used for this evaluation. 
Experiments are designed as follows.
Given three large-scale ReID datasets, there are in total six training sequences of length $2$, such as (Market$\rightarrow$Duke) and six sequences of length $3$, such as (Market$\rightarrow$Duke$\rightarrow$MSMT).
We use the baseline method described in Sec.~\ref{sec:overall_framework} and train it on all sequences separately.
After training on each dataset of every sequence, we evaluate the latest model on the first dataset of the corresponding sequence and record the performance decreases.
Both the \textbf{Data-Free} and \textbf{Replay} settings are tested.
For the Replay settings, the exemplars are selected by randomly sampling one image for each identity.
Compared to the original training sets, the size of the exemplar set for Market, Duke, and MSMT is only $5.5\%$, $4.2\%$, and $3.4\%$, respectively.
Note that in Replay settings, the old classifiers will also be updated in training.
The corresponding results are shown in Tab.~\ref{tab:results_incremental}.
To better demonstrate our improvements, we report the averaged results of the sequences that are of the same length and share the same initial dataset, {\em e.g.}, averaging the results of testing Market on the sequences Market$\rightarrow$Duke and Market$\rightarrow$MSMT.
In short, our formulation outperforms the dataset-based formulation in all experiments.
These results further demonstrate the effectiveness of our formulation.

\begin{table}[t]
\centering
\caption{
Results of ReID models on incremental learning tasks.
Each result denotes the percentage of the performance preserved on the first dataset after learning on new datasets.
$\S$ marks the Data-Free settings.
$\dag$ corresponds to the Replay settings.
}
\setlength{\tabcolsep}{1.2mm}{
\begin{tabular}{cc|cc|cc|cc}
\thickhline
\multicolumn{2}{c|}{{Testing Set}}                              & \multicolumn{2}{c|}{{Market}}                 & \multicolumn{2}{c|}{{Duke}}                   & \multicolumn{2}{c}{{MSMT}}                   \\ \hline
\multicolumn{1}{c|}{{Seq Length}}         & {   Formulation   }   & {Rank-1}  &{mAP}    & {Rank-1}  & {mAP}    & {Rank-1}  & {mAP}    \\ \hline \hline
\multicolumn{1}{c|}{{1}}                  & {$-$}         & 100\%            & 100\%          & 100\%            & 100\%             & 100\%            & 100\%           \\ \hline
\multicolumn{1}{c|}{\multirow{4}{*}{{2}}} & {Conventional$^{\S}$}  & 82.2\%          & 62.5\%          & 80.2\%                & 68.8\%          & 55.5\%               & 38.7\%          \\
\multicolumn{1}{c|}{}                            & {Ours$^{\S}$}      & 88.3\%          & 71.2\%          & 89.3\%              & 83.2\%          & 74.5\%                 & 58.9\%          \\
\multicolumn{1}{c|}{}                            & {Conventional$^\dag$} & 92.5\%          & 84.1\%          & 90.9\%             & 84.7\%          & 81.7\%              & 70.1\%          \\
\multicolumn{1}{c|}{}                            & {Ours$^\dag$}     & \textbf{95.0\%}  & \textbf{85.7\%} & \textbf{94.3\%}   & \textbf{91.1\%} & \textbf{91.6\%}   & \textbf{84.6\%} \\ \hline
\multicolumn{1}{c|}{\multirow{4}{*}{{3}}} & {Conventional$^{\S}$}  & 74.8\%         & 52.2\%          & 75.2\%          & 63.0\%          & 38.9\%                  & 24.7\%          \\
\multicolumn{1}{c|}{}                            & {Ours$^{\S}$}      & 85.8\%          & 66.0\%          & 85.8\%         & 77.4\%          & 56.6\%                  & 39.4\%          \\
\multicolumn{1}{c|}{}                            & {Conventional$^\dag$} & 86.5\%           & 74.0\%          & 84.1\%         & 76.4\%          & 74.3\%                   & 60.9\%          \\
\multicolumn{1}{c|}{}                            & {Ours$^\dag$}     & \textbf{94.4\%} & \textbf{83.1\%} & \textbf{91.5\%} & \textbf{87.6\%} & \textbf{86.4\%} & \textbf{76.0\%} \\ 
\thickhline
\end{tabular}}
\label{tab:results_incremental}
\end{table}

\begin{table}[t]
\centering
\caption{
Results of combining different normalization strategies in fully-supervised learning and direct transfer.
In this table, BN and IBN correspond to the training-set-specific normalization methods.
AdaBN adapts the dataset-wise normalization statistics. 
CBN follows our camera-based formulation and aligns each camera independently.
}
\setlength{\tabcolsep}{1.9mm}{
\begin{tabular}{c|c|cc|cc}
\thickhline
\multirow{2}{*}{{Training Method}} & \multirow{2}{*}{{Testing Method}} & \multicolumn{2}{c|}{{Duke to Duke}} & \multicolumn{2}{c}{{Duke to Market}} \\ \cline{3-6} 
                                          &                  & {Rank-1}       & {mAP}       & {Rank-1}        & {mAP}        \\ \hline \hline
{BN}              & {BN}                       & 81.5                  & 66.6               & 53.2                   & 25.1                                 \\ 
{IBN~\cite{pan2018two}}  & {IBN}               & 77.6                  & 57.0               & 61.7                   & 29.5                                \\ 
{BN}              & {AdaBN}~\cite{adabn}       & 81.2                  & 66.2               & 55.8                   & 28.1                            \\ 
{BN}              & {Our CBN}                  & 80.2                  & 63.7               & 69.5                   & 40.6                                \\ 
{Our CBN}         & {Our CBN}                  & \textbf{82.5}         & \textbf{67.3}      & \textbf{72.7}          & \textbf{43.0}               \\ \thickhline
\end{tabular}}
\label{tab:ablation_bn_variants}
\end{table}

\begin{table}[t]
\centering
\caption{
The mAP of our method on fully-supervised learning and direct transfer.
We repeat each experiment 10 times and calculate the mean and variance of all results.
}
\setlength{\tabcolsep}{4.5mm}{
\begin{tabular}{c|cc|cc}
\thickhline
\multirow{2}{*}{{\# Batches}} & \multicolumn{2}{c|}{{Market to Market}} & \multicolumn{2}{c}{{Market to Duke}} \\ \cline{2-5} 
                                     & {mean}        & {variance}       & {mean}      & {variance}    \\ \hline \hline
{1}                           & 76.29                & 0.032                   & 37.34              & 0.047                \\
{5}                           & 77.21                & 0.010                   & 38.08              & 0.017                \\
{10}                          & 77.33                & 0.007                   & 38.19              & 0.008                \\
{20}                          & 77.37                & 0.005                   & 38.18              & 0.002                \\
{50}                          & \textbf{77.39}       & \textbf{0.001}          & \textbf{38.21}     & \textbf{0.001}       \\ \thickhline
\end{tabular}}
\label{tab:ablation_num_estimation}
\end{table}

\subsection{Ablation Study}
\label{sec:ablation}

The experiments above demonstrate that our camera-based formulation boosts all the mentioned tasks.
Now, we conduct more ablation studies to validate CBN.

\noindent\textbf{Comparisons between CBN and other BN variants.}
We compare CBN with three types of BN variants.
(1) BN~\cite{ioffe2015batch} and IBN~\cite{pan2018two} correspond to the methods that use training-set-specific statistics to normalize all testing data.
(2) AdaBN~\cite{adabn} is a dataset-wise adaptation that utilizes the testing-set-wise statistics to align the entire testing set.
(3) The combination of BN and our CBN is to verify the importance of training ReID models with CBN.
As shown in Tab.~\ref{tab:ablation_bn_variants}, training and testing the ReID model with CBN achieves the best performance in both \textit{fully-supervised learning} and \textit{direct transfer}.

\noindent\textbf{Samples Required for CBN Approximation.}
We conduct experiments for approximating the camera-related statistics with different numbers of samples.
Note that if a camera contains less than the required number of images, we simply use all available images rather than duplicate them.
We repeat all experiments $10$ times and list the averaged results in Tab.~\ref{tab:ablation_num_estimation}.
As demonstrated, the performance is better and more stable when using more samples to estimate the camera-related statistics. 
Besides, results are already good enough when only utilizing very few samples,~\emph{e.g.}, $10$ mini-batches.
For the balance of simplicity and performance, we adopt $10$ mini-batches for approximation in all experiments.

\noindent\textbf{Compatibility with Different Backbones.}
Apart from ResNet~\cite{ResNet} used in the above experiments, we further evaluate the compatibility of CBN.
We embed CBN with other commonly used backbones: MobileNet V2~\cite{sandler2018mobilenetv2} and ShuffleNet V2~\cite{ma2018shufflenet}, and evaluate their performance on \textit{fully-supervised learning} and \textit{direct transfer}.
As shown in Tab.~\ref{tab:test_different_backbone}, the performance is also boosted significantly.

\begin{table}[t]
\centering
\caption{
Results of combining our camera-based formulation with different convolutional backbones.
The fully-supervised learning results are in \textit{italics}.
}
\setlength{\tabcolsep}{1.3mm}{
\begin{tabular}{c|c|c|cc|cc}
\thickhline
\multirow{2}{*}{{Backbone}}      & \multirow{2}{*}{{Training Set}} & {Testing Set}  & \multicolumn{2}{c|}{{Market}}            & \multicolumn{2}{c}{{Duke}}              \\ \cline{3-7} 
                                        &                                        & {Formulation}       & {Rank-1}        & {mAP}           & {Rank-1}        & {mAP}           \\ \hline \hline
\multirow{4}{*}{{MobileNet V2}~\cite{sandler2018mobilenetv2}}  & \multirow{2}{*}{{Market}} & {Conventional}    & \textit{87.7}          & \textit{69.2}          & 34.7                   & 18.9                   \\
                                        &                                        & {Ours}         & \textit{\textbf{89.8}} & \textit{\textbf{73.7}} & \textbf{54.4}          & \textbf{34.0}          \\ \cline{2-7} 
                                        & \multirow{2}{*}{{Duke}}         & {Conventional}     & 51.4          & 22.6          & \textit{79.8}                   & \textit{60.2}                   \\
                                        &                                        & {Ours}         & \textbf{70.7} & \textbf{39.0} & \textit{\textbf{79.9}}          & \textit{\textbf{62.4}}           \\ \hline
\multirow{4}{*}{{ShuffleNet V2}~\cite{ma2018shufflenet}} & \multirow{2}{*}{{Market}}       & {Conventional}    & \textit{82.6}          & \textit{58.4}          & 34.6        & 18.4                  \\
                                        &                                        & {Ours}         & \textit{\textbf{85.9}} & \textit{\textbf{65.8}} & \textbf{53.8}          & \textbf{33.8}          \\ \cline{2-7} 
                                        & \multirow{2}{*}{{Duke}}         & {Conventional}     & 48.1          & 20.3          & \textit{74.7}                   & \textit{52.8}                             \\
                                        &                                        & {Ours}         & \textbf{70.0} & \textbf{38.9} & \textit{\textbf{77.1}}          & \textit{\textbf{58.6}}           \\ \thickhline
\end{tabular}}
\label{tab:test_different_backbone}
\end{table}

\section{Conclusions}
In this paper, we advocate for a novel camera-based formulation for person re-identification (ReID) and present a simple yet effective solution named camera-based batch normalization. 
With only a few additional costs, our approach shrinks the gap between intra-camera learning and inter-camera learning. 
It significantly boosts the performance on multiple ReID tasks, regardless of the source of supervision, and whether the trained model is tested on the same dataset or transferred to another dataset.

Our research delivers two key messages. 
\textbf{First}, it is crucial to align \textit{all} camera-related distributions in ReID tasks, so the ReID models can enjoy better abilities to generalize across different scenarios as well as transfer across multiple datasets. 
\textbf{Second}, with the aligned distributions, we unleash the potential of intra-camera annotations, which may have been undervalued in the community. With promising performance under the weakly-supervised setting (only intra-camera annotations are available), our approach provides a practical solution for deploying ReID models in large-scale, real-world scenarios.

\section*{Acknowledgements}

This work was supported by National Science Foundation of China under grant No. 61521002.

\appendix
\section*{\LARGE{Appendix}}

\section{Camera-based Testing Scheme in Section 3.3}

In this section, we introduce the testing scheme of our camera-based formulation.
Unlike the conventional BN~\cite{ioffe2015batch}, which only calculates the statistics in the training stage and directly uses the recorded value for testing, our camera-based formulation with CBN utilizes a symmetrical approach, {\em i.e.}, estimating the camera-related statistics in both training and testing stages.

\begin{algorithm}[h]
\caption{Inference with CBN layers}
\label{alg:cbn_estimate}
\begin{algorithmic}
\STATE \textbf{Input:} a trained feature extractor $\mathbf{f\left(\cdot\right)}$, images from the testing camera set $\mathcal{C}$.
\STATE \textbf{Initialize:} grouping testing images according to their camera ID and randomly samples $N$ mini-batches from each group, denoted as $\left\{\mathbf{I}_{i}\right\}^{\left(c\right)}$
\FORALL {$c\leftarrow 1$ to $|\mathcal{C}|$}
\STATE Forward all images from $\left\{\mathbf{I}_{i}\right\}^{{\left(c\right)}}$ in $N$ mini-batches
\FORALL {CBN layers in $\mathbf{f\left(\cdot\right)}$}
\STATE Collect the corresponding mini-batch mean ${\mu}_{n}$ and variance $\sigma_{n}^\mathrm{2}$
\STATE $\hat{\mu}_{\left(c\right)} = \mathrm{accumulate}\left\{{\mu}_{1}, {\mu}_{2}, ..., {\mu}_{N}\right\}$
\STATE $\hat{\sigma}_{\left(c\right)}^\mathrm{2} = \mathrm{accumulate}\left\{\sigma_{1}^\mathrm{2},\sigma_{2}^\mathrm{2},...,\sigma_{N}^\mathrm{2}\right\}$
\STATE Inject $\hat{\mu}_{\left(c\right)}$ and $\hat{\sigma}_{\left(c\right)}^\mathrm{2}$ into the corresponding CBN layer
\ENDFOR 
\FORALL {images $\mathbf{I}^{\left(c\right)}$ from camera $c$}
\STATE Compute final features $\mathbf{f}\left(\mathbf{I}^{\left(c\right)}\right)$
\ENDFOR 
\ENDFOR 
\end{algorithmic}
\end{algorithm}

The method used in the training stage is introduced in Section 3.3.
In the testing stage, before generating the final features for each testing image, we first cluster these images according to their camera labels.
For each of these camera-related clusters, we randomly collect several unlabeled images.
Then, we group these images into mini-batches and forward them across the ReID network.
In this stage, the standardization procedure in every CBN uses mini-batch statistics, {\em i.e.}, the same procedure in training.
For each mini-batch, we collect the mini-batch mean and variance of every CBN layer.
After forwarding all related mini-batches, we approximate the overall mean and variance of each CBN layer with these mini-batch statistics using the same way in the conventional BN.
Finally, we inject our estimated results into each CBN layer and generate the final features of all images from this specific camera.
The above procedure ends when images from all testing cameras are processed.
The detailed algorithm is presented in Algorithm.~\ref{alg:cbn_estimate}.

\section{The Warm-Up Strategy in Section 4.1}

In this section, we describe the warm-up strategy for initializing fully-connected classifiers in incremental learning tasks.
Given a model that has already been trained on one or multiple ReID datasets, when fine-tuning it on a new training set, a new fully-connected classifier for classifying images from this specific dataset is required.
Since this classifier is randomly initialized, if we directly fine-tune the entire model in an end-to-end manner, this classifier will introduce lots of noises to the feature extractor and heavily damage the previously learned knowledge.
To alleviate the knowledge forgetting in the early stage of training, we warm-up the newest classifier before the formal training.
Note that in the Replay incremental learning, there could be classifiers and images that correspond to multiple training sets (the exemplar memory and the current training set).
However, in the warm-up stage of all incremental learning tasks, we only consider the latest training set and the corresponding new classifier.
The details of this warm-up strategy are presented in Algorithm~\ref{alg:warmup}.
In short, we freeze all previously learned layers and only iteratively fine-tune the new classifier on the latest training set until the loss becomes stable.
After the warm-up stage, we start to train the entire network in a conventional end-to-end manner.

\begin{algorithm}[!h]
\caption{Warm-up the latest classifier}
\label{alg:warmup}
\begin{algorithmic}
\STATE \textbf{Input:} a trained ReID model with the feature extractor $\mathbf{f\left(\cdot\right)}$, image $\mathbf{I}_{i}$ and the corresponding ID $\mathbf{y}_{i}$ from the latest training set $\mathcal{D}$
\STATE \textbf{Initialize:} freeze all trainable parameters in $\mathbf{f\left(\cdot\right)}$, randomly initialize a new classifier $\mathbf{g}\left(\cdot\right)$ for $\mathcal{D}$,
set counter $n=0$, set an empty list $\mathfrak{L}=\left[\right]$
\REPEAT 
\STATE Randomly sample a mini-batch $\left\{\mathbf{I}_{i}\right\}$ and the corresponding $\left\{\mathbf{y}_{i}\right\}$ from $\mathcal{D}$
\STATE $\mathcal{L}=\mathrm{get\_loss}\left(\mathbf{g}\left(\mathbf{f}\left(\left\{\mathbf{I}_{i}\right\}\right)\right), \left\{\mathbf{y}_{i}\right\}\right)$
\STATE Backward $\mathcal{L}$ and only update $\mathbf{g}\left(\cdot\right)$
\STATE Append $\mathcal{L}$ to $\mathfrak{L}$
\STATE Truncate $\mathfrak{L}$ and only preserve the latest $50$ items
\IF{($\mathfrak{L}$ has $50$ items) \& ($|\mathcal{L}-\mathrm{mean}\left(\mathfrak{L}\right)|<=0.1$)} \STATE $n=n+1$ \ELSE \STATE$n=0$ \ENDIF
\UNTIL{$n=5$}
\end{algorithmic}
\end{algorithm}

Our experiments show that this warm-up stage is essential for preserving the previously-learned knowledge.
Without the warm-up, the conventional BN-based method loses another 5.3\% Rank-1 accuracy on average, while our formulation loses 3.7\% on average.

\section{Exemplar Memory in Section 4.2}

\begin{algorithm}[t]
\caption{Build the exemplar memory}
\label{alg:build_exemplar}
\begin{algorithmic}
\STATE \textbf{Input:} a ReID set $\mathcal{D}$ with an identity set $\mathcal{K}$ and a camera set $\mathcal{C}$
\STATE \textbf{Output:} the exemplar memory $\mathcal{M}$ in which each identity from $\mathcal{D}$ has exactly one image
\STATE \textbf{Initialize:} create a dict $\Omega$ that records the number of already picked images from each camera.
\FORALL {identity $k$ in $\mathcal{K}$}
\STATE Collect all images that belong to the identity $k$
\STATE Collect the camera ID of the above images as $\left\{c\right\}$
\STATE Query $\Omega$ with $\left\{c\right\}$ and find the camera $c$ that has the least picked images
\STATE Randomly pick an image that simultaneously belongs to camera $c$ and identity $k$, and add it to $\mathcal{M}$
\STATE $\Omega\left[c\right]=\Omega\left[c\right]+1$
\ENDFOR 
\end{algorithmic}
\end{algorithm}

The exemplar memory is built for the Replay incremental learning task.
Its goal is to reinforce the discriminative knowledge of the previous training sets with the least amount of old images. 
In this paper, we design a straightforward approach to achieve this goal.
For each old training set, we propose a greedy algorithm that saves one image for each identity and tries to keep an equal number of images for each old camera.
The details are presented in Algorithm~\ref{alg:build_exemplar}.
With this approach, the size of the exemplar memory for Market~\cite{MARKET}, Duke~\cite{zheng2017unlabeled}, and MSMT17~\cite{PTGAN} is only $5.5\%$, $4.2\%$, and $3.4\%$ of their original training set, respectively.

\begin{figure}[h]
\centering
\includegraphics[width=0.8\textwidth]{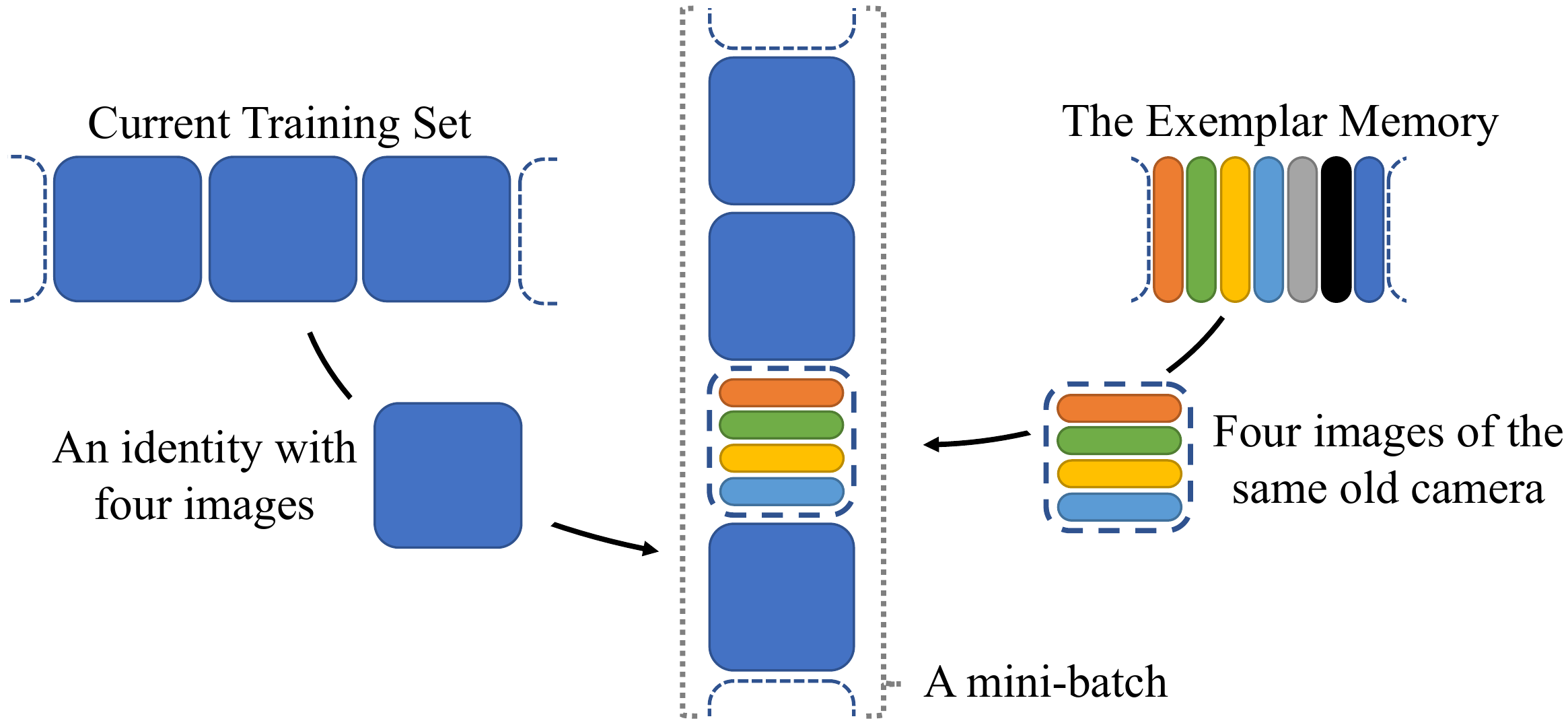}
\caption{
The demonstration of a mini-batch.
(1) A blue rectangle denotes four images of the same identity.
(2) The rectangles in other colors represent the images from the exemplar memory.
Each rectangle corresponds to one image of an old identity.
We group these exemplars according to their camera ID, and randomly fuse these groups with the data sampled from the current training set.
}
\label{fig:demo_sampler}
\end{figure}

Another thing worth noting is the way of utilizing these exemplars together with the data from the latest training set.
On the one hand, in the exemplar memory, there are only very few samples that describe the previous cameras, and each old identity only has one image.
On the other hand, as described in Section 3.3 and Section 4.1, for the latest training set, each identity has multiple images in the mini-batch, so does each camera. 
To make sure that our method can accurately approximate the CBN statistics of all previous and current cameras, we design a mixed sampling strategy.
As shown in Fig.~\ref{fig:demo_sampler}, when handling images from the latest training set, we follow the pipeline presented in Section 3.3.
When sampling identities from the exemplar memory, we cluster images from the exemplar memory and make sure that each group has four successive old images that correspond to the same old camera.
Then, these groups are randomly fused with the images sampled from the latest training set.

\section{Experiments on partially Replacing BN with CBN}

These are supplementary experiments for demonstrating the necessity of replacing \textbf{all} BN layers with CBN layers, rather than only part of them.
We go back to our baseline and divide the BN layers into six parts: the BN that appears before all residual blocks, the BN within each of the four residual stages, and the BN that appears after all blocks.
The following table summarizes the \textit{direct transfer} performance when the model trained on Duke is tested on Market.
Since the vanilla BN is below satisfaction in the \textit{direct transfer} experiments, we utilize AdaBN for adapting testing set statistics.

\begin{table}[]
\centering
\caption{
The direct transfer performance from Duke to Market.
$\checkmark$ marks the component in which all its BN layers are replaced with CBN layers.
}
\setlength{\tabcolsep}{1.9mm}{
\begin{tabular}{cccccccc}
\thickhline
First\textbf{} BN & Block 1 & Block 2 & Block 3 & Block 4 & Last BN & Rank-1        & mAP           \\ \hline \hline
       &         &         &         &         &         & 55.8          & 28.1          \\ \hline
\checkmark      &         &         &         &         &         & 60.6          & 31.6          \\ \hline
\checkmark      & \checkmark       &         &         &         &         & 61.9          & 32.9          \\ \hline
\checkmark      & \checkmark       & \checkmark       &         &         &         & 65.0          & 35.3          \\ \hline
\checkmark      & \checkmark       & \checkmark       & \checkmark       &         &         & 65.7          & 35.7          \\ \hline
\checkmark      & \checkmark       & \checkmark       & \checkmark       & \checkmark       &         & 67.3          & 37.0          \\ \hline
\checkmark      & \checkmark       & \checkmark       & \checkmark       & \checkmark       & \checkmark       & \textbf{72.7} & \textbf{43.0} \\ \thickhline
\end{tabular}
}
\end{table}

These results indicate that replacing all BN layers with CBN layers obtains the best results in the \textit{direct transfer}.
More importantly, we emphasize that only replacing part of BN layers contradicts the fundamental idea of this paper, because we believe that distribution statistics should only be collected within a camera, and all camera-related distributions should be aligned explicitly.

\bibliographystyle{splncs04}
\bibliography{egbib}
\end{document}